\documentclass{article}

\usepackage[final]{neurips_2024}

\usepackage[utf8]{inputenc} 
\usepackage[T1]{fontenc}    
\usepackage{hyperref}       
\usepackage{url}            
\usepackage{booktabs}       
\usepackage{amsfonts}       
\usepackage{nicefrac}       
\usepackage{microtype}      
\usepackage{xcolor}         
\usepackage{amsmath}
\usepackage{amsfonts}
\usepackage{amssymb}
\usepackage{graphicx}
\usepackage{subcaption}
\usepackage{makecell}
\usepackage{placeins}
\usepackage{float}
\usepackage{soul}
\soulregister{\cite}7
\soulregister{\ref}7
\usepackage{tabularx}

\title{DECRL: A Deep Evolutionary Clustering Jointed Temporal Knowledge Graph Representation Learning Approach}

\author{%
  Qian Chen, Ling Chen\thanks{Corresponding author: Ling Chen} \\
  State Key Laboratory of Blockchain and Data Security\\
  College of Computer Science and Technology\\
  Zhejiang University\\
  \texttt{\{qianchencs,lingchen\}@cs.zju.edu.cn} \\
}

\begin{document}
\maketitle

\begin{abstract}
\label{abstract}
  Temporal Knowledge Graph (TKG) representation learning aims to map temporal evolving entities and relations to embedded representations in a continuous low-dimensional vector space. However, existing approaches cannot capture the temporal evolution of high-order correlations in TKGs. To this end, we propose a \textbf{\underline{D}}eep \textbf{\underline{E}}volutionary \textbf{\underline{C}}lustering jointed temporal knowledge graph \textbf{\underline{R}}epresentation \textbf{\underline{L}}earning approach (\textbf{DECRL}). Specifically, a deep evolutionary clustering module is proposed to capture the temporal evolution of high-order correlations among entities. Furthermore, a cluster-aware unsupervised alignment mechanism is introduced to ensure the precise one-to-one alignment of soft overlapping clusters across timestamps, thereby maintaining the temporal smoothness of clusters. In addition, an implicit correlation encoder is introduced to capture latent correlations between any pair of clusters under the guidance of a global graph. Extensive experiments on seven real-world datasets demonstrate that DECRL achieves the state-of-the-art performances, outperforming the best baseline by an average of 9.53\%, 12.98\%, 10.42\%, and 14.68\% in MRR, Hits@1, Hits@3, and Hits@10, respectively.
\end{abstract}

\section{Introduction}
\label{Introduction}
Temporal Knowledge Graphs (TKGs) are collections of human temporal evolving knowledge \cite{xu2023}, which are widely utilized in various fields, e.g., information retrieval \cite{liu2018}, natural language understanding \cite{chen2019}, and recommendation systems \cite{wang2019}. A TKG represents events in the form of quadruples $(s, r, o, t)$, where $s$ and $o$ denote the subject and object entities, respectively, $r$ denotes the relation between $s$  and $o$, and $t$ represents the timestamp \cite{xu2023}. Event prediction in TKGs is an important task that predicts future events according to historical events \cite{liu2022danet}. TKG representation learning aims to map temporal evolving entities and relations to embedded representations in a continuous low-dimensional vector space. Due to the complex temporal dynamics and multi-relations within TKGs, TKG representation learning poses great challenges to the research community.

In recent years, many TKG representation learning approaches have used graph neural networks \cite{wuweathergnn} (GNNs) to model pairwise entity correlations \cite{li2022,bai2023}. Furthermore, some researchers have leveraged derived structures, e.g., communities \cite{zhang2022}, entity groups \cite{tang2023}, and hypergraphs \cite{shang2024mshyper,tang2024}, to model high-order correlations among entities, i.e., the simultaneous correlations among three or more entities.

These approaches, however, lack the capability to capture the temporal evolution of high-order correlations in TKGs. As a kind of dynamic graph, TKGs inherently feature the temporal evolution of high-order correlations. For example, within TKGs focused solely on countries, a country entity may be affiliated with different international political organizations at different timestamps, with these organizations experiencing membership adjustment over time. In addition, the membership adjustment in international political organizations does not shift suddenly. Typically, numerous events occur before the formal adjustment of their membership, gradually pushing memberships away from or drawing them closer to international political organizations. For instance, the UK’s official departure from the European Union (EU) was preceded by numerous events that incrementally estranged it from the EU.

To address the aforementioned deficiencies, a \textbf{\underline{D}}eep \textbf{\underline{E}}volutionary \textbf{\underline{C}}lustering jointed temporal knowledge graph \textbf{\underline{R}}epresentation \textbf{\underline{L}}earning approach (\textbf{DECRL}) is proposed for event prediction in TKGs. To the best of our knowledge, DECRL is the first work that integrates deep evolutionary clustering approaches into TKGs, which jointly optimizes TKG representation learning with evolutionary clustering to capture the temporal evolution of high-order correlations. Our main contributions are outlined as follows:
    \begin{itemize}
        \item We propose a deep evolutionary clustering module to capture the temporal evolution of high-order correlations among entities, where clusters represent the high-order correlations between multiple entities. Furthermore, a cluster-aware unsupervised alignment mechanism is introduced to ensure precise one-to-one alignment of soft overlapping clusters across timestamps, maintaining the temporal smoothness of clusters over successive timestamps.
        \item We propose an implicit correlation encoder to capture latent correlations between any pair of clusters, which defines the interaction intensities between clusters to form a cluster graph. In addition, a global graph, constructed from all events of training set, is introduced to guide the assignment of different interaction intensities to different cluster pairs. 
        \item We evaluate DECRL on seven real-world datasets. The experimental results demonstrate that DECRL achieves the state-of-the-art (SOTA) performance. It outperforms the best baseline by an average of 9.53\%, 12.98\%, 10.42\%, and 14.68\% in MRR, Hits@1, Hits@3, and Hits@10, respectively.
    \end{itemize}

\section{Related work}
\label{related_work}
\subsection{TKG representation learning approach}
GNNs and variants of recurrent neural networks (RNNs) are commonly integrated to model graph structural information and temporal dependency within TKGs, respectively \cite{bai2023,wu2020,li2021temporal,li2022}. For example, TeMP \cite{wu2020} and RE-GCN \cite{li2021temporal} both utilize relation-aware GCN \cite{schlichtkrull2018} to model the influence of neighbor entities and apply GRUs to model the temporal dependency. TiRGN \cite{li2022} employs multi-relational GNNs to model graph structural information, and utilizes GRUs to capture the temporal dependency across sequential timestamps. However, these approaches often resort to stacking multiple layers to model the influence of distant neighbors, which may lead to the over-smoothing problem.

Recent research advancements have introduced paths \cite{li2021search,han2020,liu2022TLogic} to enhance the modeling capability of the latent pairwise correlations between entities. For example, xERTE \cite{han2020} utilizes a time-aware graph attention mechanism to obtain and exploit local subgraphs. TLogic \cite{liu2022TLogic} employs a time-based random walk strategy to acquire logical paths. Furthermore, some researchers have leveraged derived structures, e.g., communities \cite{zhang2022}, entity groups \cite{tang2023}, and hypergraphs \cite{tang2024}, to model high-order correlations among entities. For example, EvoExplore \cite{zhang2022} utilizes a temporal point process enhanced by a hierarchical attention mechanism to establish dynamic communities for modeling latent correlations among entities.  DHyper \cite{tang2024} utilizes hypergraph neural networks to model high-order correlations among entities and among relations.

While the promising results of introducing derived structures, existing approaches lack the capability to capture the temporal evolution of high-order correlations in TKGs. To address this gap, we propose DECRL, which is the first work to integrate deep evolutionary clustering approaches into TKGs. DECRL jointly optimizes TKG representation learning with evolutionary clustering to capture the temporal evolution of high-order correlations.

\subsection{Deep evolutionary clustering approach}
Existing deep evolutionary clustering approaches employ different strategies for discovering stable clusters \cite{folino2013,ma2017,yang2023,ma2020,you2021,yao2021}. For example, DYNMOGA \cite{folino2013} employs a deep evolutionary clustering approach designed to identify evolving communities within dynamic networks, which treats the process as a multi-objective optimization problem aimed at cluster stability. sE-NMF \cite{ma2017} implements linear fusion of adjacency matrices across timestamps to enhance the temporal stability of clusters. Both DNETC \cite{yang2023} and CDNE \cite{ma2020} use constructed temporal smoothness loss functions to ensure a consistent and stable evolution of clustering results through successive timestamps. TRNNGCN \cite{yao2021} introduces a decay matrix to assess the influence of historical clustering results, thereby stabilizing the current clustering configuration.

These approaches straightforwardly incorporate prior clustering results with current timestamp results over time under the presumption that clustering results are relatively stable across different timestamps. However, the nuances between similar clusters are difficult to recognize in this way. In TKG representation learning, it is necessary to discern the subtle differences in high-order correlations. To bridge this gap, in our work, a cluster-aware unsupervised alignment mechanism is introduced, which ensures the precise one-to-one alignment of soft overlapping clusters across timestamps, thereby maintaining the temporal smoothness of clusters over successive timestamps.

\vspace{-5pt}
\section{Preliminaries}
\label{preliminaries}
\vspace{-5pt}
\textbf{Definition 1 (TKG).} A TKG is a set of events formalized as ${\mathcal{G} = \{(s, r, o, t) \mid s, o \in \mathcal{E}, r \in \mathcal{R}, t \in \mathcal{T}\}}$, where $\mathcal{E}$, $\mathcal{R}$, and $\mathcal{T}$ denote the set of entities, relations, and timestamps, respectively. $\mathcal{G}^t$ denotes the set of events at timestamp \textit{t}.

\textbf{Definition 2 (Entity Graph).} The entity graph is constructed based on $\mathcal{G}^t$, which is a multi-relational graph, and can be denoted as $G_{\text{e}}^t = (V_{\text{e}}^t, E_{\text{e}}^t)$, where $V_{\text{e}}^t$ and $E_{\text{e}}^t$ denote the set of nodes and edges of the entity graph at timestamp $t$, respectively. The nodes in $V_{\text{e}}^t$ represent entities, while the edges in $E_{\text{e}}^t$ represent relations between entities at timestamp $t$.

\textbf{Definition 3 (Cluster Graph).} The cluster graph is constructed based on $G_\text{e}^t$, which is a fully connected graph, and can be denoted as $G_{\text{c}}^t = (V_{\text{c}}^t, E_{\text{c}}^t)$, where $V_{\text{c}}^t$ and $E_{\text{c}}^t$ denote the set of nodes and edges of the cluster graph at timestamp $t$, respectively. The nodes and edges of cluster graphs represent clusters and the latent correlations between them, respectively, where each cluster represents the high-order correlations among entities.

\textbf{Task (Event Prediction).} Given a query $(s, ?, o, t)$, the event prediction task in TKGs aims to predict the conditional probability of all relations with the subject entity $s$, the object entity $o$, and the historical event sets $\mathcal{G}^{1:T-1}$, which can be denoted as $p(\hat{\boldsymbol{r}}|s, o, \mathcal{G}^{1:T-1})$, where $T$ denotes the number of historical timestamps.
\begin{figure}[]
    \centering
    \includegraphics[width=0.75\linewidth]{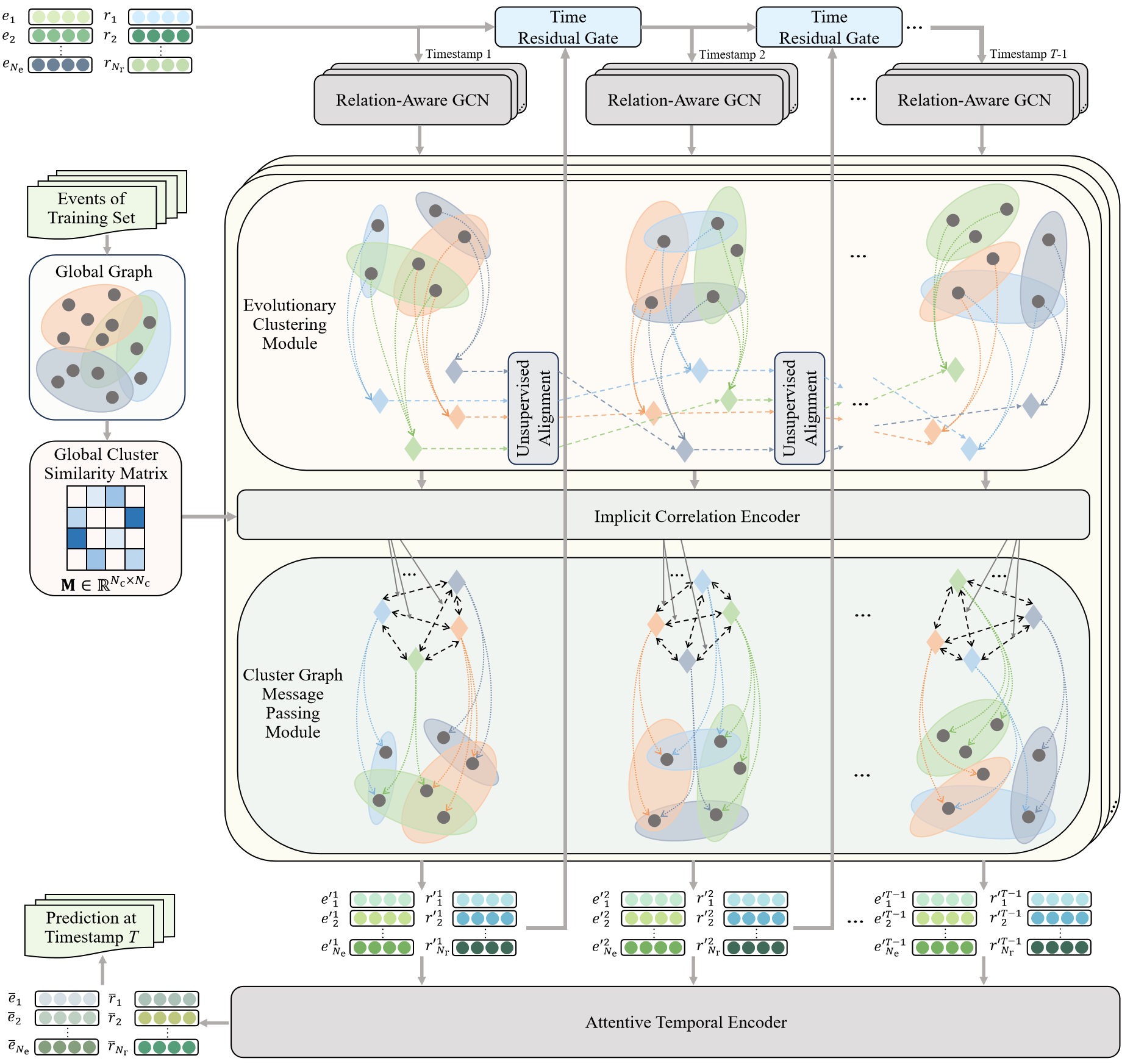}
    \caption{The framework of DECRL.}
    \label{fig:whole_DECRL}
\end{figure}

\section{Methodology}
\label{Methodology}
\vspace{-5pt}
\subsection{Overview}
The proposed DECRL approach is illustrated in Figure \ref{fig:whole_DECRL}. At each timestamp, entity and relation representations updated by the cluster graph message passing module are merged with input representations from the previous timestamp using a time residual gate, serving as the input for the current timestamp. The multi-relational interactions among entities are modeled by the relation-aware GCN. The deep evolutionary clustering module captures the temporal evolution of high-order correlations among entities, maintaining the temporal smoothness of clusters over successive timestamps through an unsupervised alignment mechanism. The implicit correlation encoder captures the latent correlations between any pair of clusters, enabling message passing within the cluster graph to update entity representations. Finally, the attentive temporal encoder captures temporal dependencies among entity and relation representations across timestamps, integrating them with temporal information for future event prediction. The computational complexity of DECRL can be found in Appendix \ref{complexity}.

\subsection{Evolutionary clustering module}
\label{sec:evoclumodule}
The entity graph $G_{\text{e}}^t = (V_{\text{e}}^t, E_{\text{e}}^t)$ is constructed based on the historical events at timestamp $t$, which can be used to capture the structural dependency among concurrent events. Since the entity graph is a multi-relational graph, relation-aware GCN \cite{schlichtkrull2018} is utilized to model the structural dependency, which is formulated as:
\begin{equation}
\boldsymbol{h}_\text{o}^{t,l+1}=\mathrm{RReLU}\left(\frac1{d_\text{o}}\sum_{(s,r),\exists(s,r,o)\in\mathcal{G}^t}\boldsymbol{W}_1(\boldsymbol{h}_\text{s}^{t,l}+\boldsymbol{h}_\text{r}^{t,l})+\boldsymbol{W}_2\boldsymbol{h}_\text{o}^{t,l}\right)
\label{eq:rgcn}
\end{equation}
where $\boldsymbol{h}_\text{s}^{t,l}$, $\boldsymbol{h}_\text{o}^{t,l}$, and $\boldsymbol{h}_\text{r}^{t,l}$ denote the $l^{th}$ layer representations of subject entity $s$, object entity $o$, and relation $r$ at timestamp $t$, respectively. $\boldsymbol{W}_1$ and $\boldsymbol{W}_2$ are learnable parameters for aggregating neighbor entity representations and self-loop, respectively. ${d_\text{o}}$ denotes the in-degree of object entity $o$. Initially, entities are assigned random representations based on their in-degree, such that entities with the same in-degree have the same initial representations, thereby efficiently accelerating the update process of entity representations. Relations are given with the random initial representations. For simplicity, the subscript $l$ of variables is omitted in the following sections without causing ambiguity.

The representation of the relation $r$ at timestamp $t$ is influenced by the $r$-related entity representations at timestamp $t$ and its representation at timestamp $t-1$. The updating of the relation representation is formulated as $\boldsymbol{h}_\text{r}^t=\mathrm{pooling}[\boldsymbol{e}^{t}_{\mathcal{V}^t_\text{r}};\boldsymbol{h}_\text{r}^{t-1}]$,
where $[;]$ denotes the concatenation operation. $\mathrm{pooling}$ denotes the mean pooling operation. $\boldsymbol{e}^{t}_{\mathcal{V}^t_\text{r}}$ denotes $r$-related entity representations at timestamp $t$, where $\mathcal{V}^t_\text{r}=\{i|(i,r,o,t)\,or\,(s,r,i,t)\in\mathcal{G}^t \}$.

A country entity may be affiliated with different international political organizations at different timestamps within TKGs. Thus, the fuzzy c-means clustering method \cite{pu2024} is utilized to obtain the soft overlapping clusters, which outputs the membership matrix between entities and clusters by optimizing the object function as follows:
\begin{equation}
    J=\sum_{i=1}^{N_\text{e}}\sum_{j=1}^{N_\text{c}}{u}_{i,j}^{t,m}\left(1-\frac{\langle \boldsymbol{e}_i^t,\boldsymbol{\mu}_j^t\rangle}{\|\boldsymbol{e}_i^t\|\|\boldsymbol{\mu}_j^t\|}\right),s.t.\sum_{j=1}^{N_\text{c}}{u}_{i,j}^{t,m}=1, 0<\sum_{i=1}^{N_\text{e}}{u}_{i,j}^{t,m}<N_\text{e}
\label{eq:cmeans}
\end{equation}
where ${N_\text{e}}$ and ${N_\text{c}}$ are the number of entities and clusters, respectively. ${u}_{i,j}^t$ denotes the membership degree of the entity representation $\boldsymbol{e}_i^t$ to the cluster centroid $\boldsymbol{\mu}_j^t$ at timestamp $t$. $m$ denotes the fuzzy smoothing hyper-parameter, which is typically set to a value greater than $1$. $\langle\cdot\rangle$ denotes the dot product. $\|\cdot\|$ denotes the norm of a vector. Then, the representation of clusters is formulated as:
\begin{equation}
\begin{aligned}
\boldsymbol{c}_j^t=\sum_{\boldsymbol{e}_i^t\in V_{\text{e}}^t}{u}_{i,j}^t\boldsymbol{e}_i^t,
& \quad
    {u}_{i,j}^t=\frac{\left(\frac{\langle \boldsymbol{e}_i^t,\boldsymbol{\mu}_j^t\rangle}{\|\boldsymbol{e}_i^t\|\|\boldsymbol{\mu}_j^t\|}\right)^{\frac1{m-1}}}{\sum\limits_{k=1}^{N_\text{c}}\left(\frac{\langle \boldsymbol{e}_i^t,\boldsymbol{\mu}_k^t\rangle}{\|\boldsymbol{e}_i^t\|\|\boldsymbol{\mu}_k^t\|}\right)^{\frac1{m-1}}}
\end{aligned}
\end{equation}
where $\boldsymbol{c}_j^t$ denotes the representation of cluster $j$ at timestamp $t$, weighted by the membership degree of different entities to the cluster.

The maximum weight matching can be found in polynomial time using the Hungarian algorithm \cite{kuhn1955}. Herein, a cluster-aware Hungarian matching algorithm is introduced to ensure a smooth one-to-one alignment of clusters across timestamps. The cluster-aware Hungarian matching algorithm quantifies the similarity between clusters of consecutive timestamps, i.e., $t-1$ and $t$, which creates an affinity matrix $A$, where each element ${a}_{j,k}$ represents the similarity between the cluster $j$ at timestamp $t-1$ and the cluster $k$ at timestamp $t$:
\begin{equation}
    {a}_{j,k}=\mathrm{cos}(\boldsymbol{c}_j^{t-1},\boldsymbol{c}_k^t)
\end{equation}

With the affinity matrix established, the cluster-aware Hungarian algorithm seeks an optimal assignment that maximizes the sum of similarities across matched clusters. It is formulated as an optimization problem: $\max_\pi\sum_{j=1}^{N_\text{c}}{a}_{j,\pi(j)}$, where $\pi$ is the permutation function representing the one-to-one alignment between clusters of consecutive timestamps. Therefore, $\pi(j)$ denotes the index of the cluster at timestamp $t$ matched with the cluster $j$ at timestamp $t-1$.

Subsequently, the fused cluster representations are computed by taking a weighted combination of the cluster representations at consecutive timestamps, which is formulated as:
\begin{equation}
    \boldsymbol{c}_j^t=\beta \boldsymbol{c}_j^{t-1}+(1-\beta)\boldsymbol{c}_{\pi(j)}^t
\end{equation}
where $\beta$ is the fusion weight, which indicates the relative contribution of each cluster representation at consecutive timestamps to the fused cluster representation.

What's more, it is essential to ensure the temporal smoothness of these cluster representations across consecutive timestamps. To this end, a temporal smoothness loss is introduced to penalize significant deviations of cluster representations across consecutive timestamps, which is formulated as:
\begin{equation}
    \mathcal{L}_{\mathrm{temporal}}=\sum_{t=1}^{T-1}\sum_{j=1}^{N_\text{c}}\left(1-\frac{\langle \boldsymbol{c}_j^{t-1},\boldsymbol{c}_j^t\rangle}{\|\boldsymbol{c}_j^{t-1}\|\|\boldsymbol{c}_j^t\|}\right)
\end{equation}

\subsection{Cluster graph message passing module}
Due to the intricate relationships and alliances formed among international political organizations, it is crucial to capture the latent correlations between clusters for event prediction. Herein, an implicit correlation encoder is proposed to capture the latent correlations between any pair of clusters. The cluster graph is designed to be a fully connected graph to avoid missing any relevant information between clusters. The representations of the latent correlations are formulated as:
\begin{equation}    \boldsymbol{s}_{i,j}^t=\mathrm{ReLU}\left(\varphi(\boldsymbol{c}_i^t,\boldsymbol{c}_j^t)\right)
\label{eq:latentcorr}
\end{equation}
where $\boldsymbol{c}_i^t$ and $\boldsymbol{c}_j^t$ are representations of the cluster $i$ and $j$ at timestamp $t$, respectively. $\varphi$ is the transformation function, which is implemented by the multi-layer perceptron.

It is worth noting that clusters exhibit varying degrees of interaction, characterized by different intensities of latent correlations. Thus, we quantify the intensity of latent correlations between clusters, which is formulated as:
\begin{equation}
    q_{i,j}^t=\sigma\left(\mathrm{Conv}(\boldsymbol{s}_{i,j}^t)\right)
\end{equation}
where $\sigma$ is the sigmoid function, which ensures the resulting correlation intensity bounded between $0$ and $1$. $\mathrm{Conv}(\cdot)$ represents the convolution operation.

The intensity of the latent correlations between clusters varies in the short term and intensifies over time; a higher frequency of interactions indicates a stronger latent correlation. To this end, we construct a global graph, which is a static graph composed of all events from the training set. The spectral clustering approach is then applied to this global graph to obtain global clusters $\boldsymbol{c}_{i}^\mathrm{global}$. The similarity between global clusters is used to enhance the intensity of latent correlations between cluster pairs, which is formulated as:
\begin{equation}
\begin{aligned}
    \widehat{q}_{i,j}^t=q_{i,j}^t\times{m}_{i,j}^t,
& \quad
    {m}_{i,j}^t=\frac{\langle \boldsymbol{c}_{\pi^t(i)}^\mathrm{global},\boldsymbol{c}_{\pi^t(j)}^\mathrm{global}\rangle}{\|\boldsymbol{c}_{\pi^t(i)}^\mathrm{global}\|\|\boldsymbol{c}_{\pi^t(j)}^\mathrm{global}\|}
\end{aligned}
\label{eq:gobalenhancestrength}
\end{equation}
where $\boldsymbol{c}_{\pi^t(i)}^\mathrm{global}$ and $\boldsymbol{c}_{\pi^t(j)}^\mathrm{global}$ are global clusters matched with the cluster $i$ and $j$ at timestamp $t$, respectively. ${m}_{i,j}^t$ is the similarity between the global cluster $i$ and $j$ at timestamp $t$. $\widehat{q}_{i,j}^t$ is the enhanced intensity of latent correlations, which also integrates structural information from the global graph.

In the message passing process of the cluster graph, the stronger the intensity of the latent correlation, the more critical the latent correlation is, and vice versa. The aggregation operation is formulated as:
\begin{equation}
    \boldsymbol{v}_{c_i}^t=\sum_{i\neq j}\widehat{q}_{i,j}^t\boldsymbol{s}_{i,j}^t
\end{equation}
where $\widehat{q}_{i,j}^t$ and $\boldsymbol{s}_{i,j}^t$ denote the enhanced intensity and the representation of latent correlation calculated by Equation \ref{eq:gobalenhancestrength} and Equation \ref{eq:latentcorr}, respectively. Then, the representation of clusters is updated through:
\begin{equation}
    \boldsymbol{\widehat{c}}_i^t=\varphi(\boldsymbol{v}_{c_i}^t,\boldsymbol{c}_i^t)
\label{eq:updatedcluster}
\end{equation}

To implement information transfer from clusters to individual entities, we use the membership matrix from Section \ref{sec:evoclumodule}. Entity representations are updated based on their associated cluster representations, which are formulated as $\boldsymbol{e}_i^{\prime t}=\sum_{j=1}^{N_\text{c}}\boldsymbol{u}_{i,j}^t \boldsymbol{\widehat{c}}_j^t$, where ${N_\text{c}}$ denotes the number of clusters.

\subsection{Time residual gate}
The time residual gate is introduced to combine updated entity and relation representations with input representations of the current timestamp through a weighted mechanism. This approach preserves inherent characteristics while capturing the temporal evolution of entities and relations. For simplicity, the subscripts $i$ and $j$ of variables are omitted in the following sections without causing ambiguity. The final updated representations at timestamp $t$ are formulated as:
\begin{equation}
\begin{aligned}
    \boldsymbol{H}^t=\boldsymbol{X}^t\otimes \boldsymbol{H}_\theta^t+(1-\boldsymbol{X}^t)\otimes \boldsymbol{H}^{t-1},
& \quad
\boldsymbol{X}^t=\sigma(\boldsymbol{W}_3\boldsymbol{H}^{t-1}+\boldsymbol{b})
\end{aligned}
\end{equation}
where $\boldsymbol{H}_\theta^t$ denotes the entity or relation representations updated by the cluster graph message passing module at timestamp $t$. $\boldsymbol{H}^{t-1}$ denotes the output of entity or relation representations at timestamp $t-1$, and also is the input of timestamp $t$. The time residual gate, i.e., $\boldsymbol{X}^t$, determines the inherent characteristics to be preserved, where $\sigma$ is the sigmoid function. $\boldsymbol{W}_3$ and $\boldsymbol{b}$ are learnable parameters.

\subsection{Attentive temporal encoder}
In the context of sequence modeling, the attentive temporal encoder is introduced to capture the temporal dependency among the final updated representations across timestamps. The position-enhanced representations of entity and relation are formulated as:
\begin{equation}
\begin{aligned}
\boldsymbol{z}^t=[\boldsymbol{H}^t; \Phi(t)], & \quad     \Phi(t)=\sqrt{\frac1d}[\cos(\omega_1\tau^t),\cos(\omega_2\tau^t),...,\cos(\omega_d\tau^t)]
\end{aligned}
\end{equation}
where $\boldsymbol{H}^t$ is the final updated representation of entity or relation at timestamp $t$. $[;]$ denotes the concatenation operation. $\Phi(t)$ is the time position encoder. $d$ denotes the value of the representation dimension. $\omega_1$ to $\omega_d$ are learnable parameters. $\tau^t$ is the timestamp. Then, the temporal dependency is captured by a position-enhanced self-attention mechanism based on representation sequence $\boldsymbol{z}^{1:T-1}=\{\boldsymbol{z}^{1},\boldsymbol{z}^{2},...,\boldsymbol{z}^{T-1}\}$. For simplicity, the subscript $t$ of variables is omitted in the following sections. The integrated entity or relation representation is formulated as:
\begin{equation}
\begin{aligned}
\beta_{m,n}=\frac{\langle\boldsymbol{W}_q\boldsymbol{z}_m,\boldsymbol{W}_k\boldsymbol{z}_n\rangle}{\sqrt{d}},
& \quad
    \alpha_{m,n}=\frac{\exp(\beta_{m,n})}{\sum\limits_{i=1}^{T-1}\exp(\beta_{m,i})},
& \quad
    \overline{\boldsymbol{h}}_m=\sum_{n=1}^{T-1}{\alpha}_{m,n}\boldsymbol{z}_{m,n}
\end{aligned}
\end{equation}
where $\boldsymbol{W}_q$ and $\boldsymbol{W}_k$ are learnable parameters. $\alpha_{m,n}$ is the attention weight. $\overline{\boldsymbol{h}}_m$ is the integrated representation of entity or relation, which integrates the temporal information.

\subsection{Event prediction}
ConvTransE \cite{shang2019} is employed as the decoder of DECRL, which predicts the probability of each relation between an entity pair, which is formulated as:
\begin{equation}
    p(\hat{\boldsymbol{r}}|s, o, \mathcal{G}^{1:T-1})=\sigma(\boldsymbol{H}_\mathrm{r}\text{ConvTransE}(\bar{\boldsymbol{e}}_\text{s},\bar{\boldsymbol{e}}_\text{o}))
\label{eq:ouputp}
\end{equation}
where $\hat{\boldsymbol{r}}$ is the probability vector of relations. $\sigma$ is the sigmoid function. $\boldsymbol{H}_\text{r}$ is the relation representation matrix, each row of which corresponds to an integrated relation representation $\overline{\boldsymbol{r}}$. $\mathrm{ConvTransE}(\cdot)$ contains a one-dimensional convolution layer and a fully connected layer.

The event prediction training objective is minimizing the cross-entropy loss, which is formulated as:
\begin{equation}
    \mathcal{L}_{\mathrm{TKG}}=-\frac{1}{S}\sum_{i=1}^{S}\sum_{j=1}^{N_\text{r}}\{y_{i,j}\log p_{i,j}+(1-y_{i,j})\log(1-p_{i,j})\}
\end{equation}
where $S$ and $N_\text{r}$ denote the numbers of samples in the training set and relations, respectively. $y_{i,j}$ denotes the label of relation $j$ for sample $i$, of which the element is $1$ if the event occurs, otherwise $0$. $p_{i,j}$ is the predicted probability of relation $j$ for sample $i$, calculated by Equation \ref{eq:ouputp}.

Finally, the total loss for DECRL is formulated as:
\begin{equation}
\mathcal{L}=(1-\lambda)\mathcal{L}_{\mathrm{TKG}}+\lambda\mathcal{L}_{\mathrm{temporal}}
\label{eq:loss}
\end{equation}
where $\lambda$ is a hyper-parameter that controls the trade-off between event prediction and temporal smoothness, which is bounded between $0$ and $1$.

\section{Experiments}
In this section, we evaluate the performance of DECRL through comprehensive experiments. We first describe the datasets and experimental settings, followed by a comparison with 12 SOTA TKG representation learning approaches. Next, we present an ablation study and hyper-parameter sensitivity analysis (see Appendix \ref{sensitivity}). Finally, we offer case studies showcasing the effectiveness of DECRL.

\subsection{Datasets and experimental settings}
We evaluate DECRL on seven real-world datasets: ICEWS14 \cite{trivedi2017}, ICEWS18 \cite{jin2020}, ICEWS14C \cite{tang2024}, ICEWS18C \cite{tang2024}, GDELT \cite{leetaru2013gdelt}, WIKI \cite{leblay2018deriving}, and YAGO \cite{jin2020}. The ICEWS14 and ICEWS18 datasets span January 1, 2014, to December 31, 2014, and January 1, 2018, to October 31, 2018, respectively. We derive ICEWS14C and ICEWS18C datasets by filtering the original datasets to focus exclusively on events involving countries. The GDELT dataset spans January 1, 2018, to January 31, 2018. The WIKI and YAGO datasets are subsets of the Wikipedia history and YAGO3 \cite{mahdisoltani2013yago3}, respectively. Dataset statistics and experimental settings are detailed in Appendices \ref{stat_dataset} and \ref{exp_setting}.

\subsection{Comparison with baselines}
\label{comparison_baselines}

\begin{table}[]
\vspace{-10pt}
\caption{The performance of DECRL and the compared approaches on ICEWS14 and ICEWS14C}
\label{Comparison14}
\centering
\renewcommand{\arraystretch}{0.9}
\setlength{\tabcolsep}{4pt}
\setlength{\abovetopsep}{0pt}
\setlength{\belowbottomsep}{0pt}
\resizebox{\linewidth}{!}{
\begin{tabular}{lllllllll}
\toprule
                      & \multicolumn{4}{c}{ICEWS14}       & \multicolumn{4}{c}{ICEWS14C}      \\ \cmidrule{2-9} 
Approach              & MRR   & Hits@1 & Hits@3 & Hits@10 & MRR   & Hits@1 & Hits@3 & Hits@10 \\ \midrule
TTransE (WWW 2018)    & 23.79* & 14.24*  & 29.17*  & 34.56*   & 11.79* & 13.24*  & 19.97*  & 24.88*   \\
HyTE (EMNLP 2018)     & 25.12* & 18.15*  & 30.15*  & 45.37*   & 22.17* & 18.15*  & 27.28*  & 35.37*   \\ \midrule
RE-NET (EMNLP 2020)   & 45.77* & 37.98*  & 49.07*  & 58.87*   & 43.27* & 36.97*  & 47.08*  & 55.19*   \\
Glean (KDD 2020)      & 42.20* & 36.86*  & 47.68*  & 52.39*   & 40.24* & 34.62*  & 45.48*  & 50.09*   \\
TeMP (EMNLP 2020)     & 46.04* & 39.07*  & 49.84*  & 59.74*   & 44.17* & 37.37*  & 47.78*  & 55.66*   \\
RE-GCN (SIGIR 2021)   & 45.56* & 38.09*  & 50.37*  & 62.44*   & 41.76* & 36.67*  & 45.37*  & 51.74*   \\
DACHA (TKDD 2022)     & 45.44* & 37.88*  & 49.47*  & 58.69*   & 44.26* & 37.59*  & 44.18*  & 53.19*   \\
TiRGN (IJCAI 2022)    & 46.07* & 39.83*  & 52.17*  & 63.95*   & 44.73* & 38.13*  & 49.77*  & 60.91*   \\ \midrule
TITer (EMNLP 2021)    & 46.12* & 39.08*  & 50.76*  & 60.39*   & 44.86* & 39.37*  & 48.84*  & 55.79*   \\
EvoExplore (KBS 2022) & 47.71* & 40.68*  & 52.37*  & 65.94*   & 49.77* & 40.12*  & 54.37*  & 65.83*   \\
GTRL (TKDE 2023)      & 46.25* & 40.11*  & 51.09*  & 65.79*   & 50.95* & 40.31*  & 52.09*  & 64.89*   \\
DHyper (TOIS 2024)    &\underline{56.15}* &\underline{43.76}*  &\underline{65.46}*  &\underline{85.89}*   &\underline{54.16}* &\underline{41.45}*  &\underline{62.03}*  &\underline{75.35}*   \\
\textbf{DECRL}      &\textbf{62.61}  &\textbf{48.73}   &\textbf{70.57}   &\textbf{93.03}    &\textbf{58.55} &\textbf{44.62}  &\textbf{66.52}  &\textbf{82.06}     \\ \midrule
\textbf{Improvement}  &\textbf{11.50\%} &\textbf{11.36\%} &\textbf{7.81\%}  &\textbf{8.31\%}   &\textbf{8.11\%}  &\textbf{7.65\%}   &\textbf{7.24\%}   &\textbf{8.91\%}   \\ \bottomrule
\end{tabular}
}
\vspace{-5pt}
\end{table}

\begin{table}[]
\caption{The performance of DECRL and the compared approaches on ICEWS18 and ICEWS18C}
\label{Comparison18}
\centering
\renewcommand{\arraystretch}{0.9}
\setlength{\tabcolsep}{4pt}
\setlength{\abovetopsep}{0pt}
\setlength{\belowbottomsep}{0pt}
\resizebox{\linewidth}{!}{
\begin{tabular}{lllllllll}
\toprule
                      & \multicolumn{4}{c}{ICEWS18}       & \multicolumn{4}{c}{ICEWS18C}      \\ \cmidrule{2-9} 
Approach              & MRR   & Hits@1 & Hits@3 & Hits@10 & MRR   & Hits@1 & Hits@3 & Hits@10 \\ \midrule
TTransE (WWW 2018)    & 11.96* & 13.97*  & 12.79*  & 24.33*   & 9.84*  & 10.29*  & 11.04*  & 18.89*   \\
HyTE (EMNLP 2018)     & 21.85* & 16.86*  & 25.64*  & 41.86*   & 22.23* & 16.27*  & 25.68*  & 33.39*   \\ \midrule
RE-NET (EMNLP 2020)   & 42.25* & 33.81*  & 44.98*  & 52.72*   & 41.05* & 32.87*  & 42.78*  & 50.43*   \\
Glean (KDD 2020)      & 37.11* & 34.15*  & 42.56*  & 47.35*   & 35.58* & 32.26*  & 40.44*  & 46.49*   \\
TeMP (EMNLP 2020)     & 43.24* & 38.77*  & 45.04*  & 55.94*   & 43.08* & 36.07*  & 43.18*  & 53.03*   \\
RE-GCN (SIGIR 2021)   & 41.56* & 37.59*  & 44.34*  & 57.42*   & 40.27* & 36.35*  & 41.75*  & 49.25*   \\
DACHA (TKDD 2022)     & 43.87* & 37.11*  & 47.47*  & 57.69*   & 40.11* & 36.11*  & 46.17*  & 52.37*   \\
TiRGN (IJCAI 2022)    & 44.27* & 38.13*  & 50.66*  & 62.90*   & 43.57* & 37.23*  & 47.67*  & 54.44*   \\ \midrule
TITer (EMNLP 2021)    & 45.44* & 39.78*  & 48.77*  & 58.73*   & 44.07* & 38.85*  & 46.44*  & 49.79*   \\
EvoExplore (KBS 2022) & 46.65* & 40.05*  & 50.07*  & 58.35*   & 47.33* & 38.96*  & 49.37*  & 56.15*   \\
GTRL (TKDE 2023)      & 46.35* & 40.95*  & 51.19*  & 60.18*   & 49.33* & 40.15*  & 53.39*  & 60.74*   \\
DHyper (TOIS 2024)    &\underline{54.22}* &\underline{42.16}*  &\underline{63.26}*  &\underline{75.38}*   &\underline{52.11}* &\underline{41.04}*  &\underline{60.03}*  &\underline{73.22}*   \\
\textbf{DECRL}      &\textbf{63.30}  &\textbf{50.13}  &\textbf{70.72}   &\textbf{90.82}    &\textbf{61.37}       &\textbf{46.28}        &\textbf{67.01}        &\textbf{86.79}         \\ \midrule
\textbf{Improvement}   &\textbf{16.75\%} &\textbf{18.90\%}  &\textbf{11.79\%}  &\textbf{20.48\%}   &\textbf{17.77\%}       &\textbf{12.76\%}        &\textbf{11.63\%}         &\textbf{18.53\%}          \\ \bottomrule
\end{tabular}
}
\vspace{-5pt}
\end{table}

\begin{table}[]
\caption{The performance of DECRL and the compared approaches on GDELT. “OOM” and “TLE” indicate out of memory and a single epoch exceeded 24 hours}
\label{Comparisongdelt}
\centering
\renewcommand{\arraystretch}{0.9}
\setlength{\tabcolsep}{4pt}
{\fontsize{9}{10}\selectfont
\begin{tabular}{lllll}
\toprule
Approach              & MRR   & Hits@1 & Hits@3 & Hits@10 \\ \midrule
TTransE (WWW 2018)    & 8.62* & 7.73*  & 11.03*  & 23.34*   \\
HyTE (EMNLP 2018)     & 10.63* & 8.39*   & 14.23*  & 28.79*   \\ \midrule
RE-NET (EMNLP 2020)   & 17.55* & 11.73*  & 18.14*  & 35.52*   \\
Glean (KDD 2020)      & 15.60* & 10.35*  & 17.61*  & 37.40*   \\
TeMP (EMNLP 2020)     & 19.19* & 11.07*  & 19.84*  & 40.52*   \\
RE-GCN (SIGIR 2021)   & 20.84* & 10.80*  & 21.09*  & 43.65*   \\
DACHA (TKDD 2022)     & 21.91* & 11.27*  & 17.49*  & 47.13*   \\
TiRGN (IJCAI 2022)    &\underline{24.61}* &\underline{13.78}*  &\underline{25.66}* & 49.02*   \\ \midrule
TITer (EMNLP 2021)    & TLE    & TLE     & TLE     & TLE   \\
EvoExplore (KBS 2022) & 18.53* & 10.74*  & 19.45*  & 42.07*   \\
GTRL (TKDE 2023)      & 22.44* & 12.48*  & 18.03*  &\underline{50.82}*   \\
DHyper (TOIS 2024)    &OOM     &OOM      &OOM      &OOM   \\
\textbf{DECRL}      &\textbf{27.58}  &\textbf{15.74}   &\textbf{29.16}   &\textbf{59.54}    \\ \midrule
\textbf{Improvement}  &\textbf{12.07\%} &\textbf{14.22\%} &\textbf{13.64\%}  &\textbf{17.16\%}   \\ \bottomrule
\end{tabular}
}
\vspace{-5pt}
\end{table}

\begin{table}[]
\caption{The performance of DECRL and the compared approaches on WIKI and YAGO with MRR}
\label{Comparisonwikiyago}
\centering
\renewcommand{\arraystretch}{0.9}
\setlength{\tabcolsep}{4pt}
{\fontsize{9}{10}\selectfont
\begin{tabular}{lll}
\toprule
Approach   & WIKI & YAGO \\ 
\midrule
TiRGN (IJCAI 2022)      & 99.04 & 99.30 \\
DHyper (TOIS 2024)      & \underline{99.38} & \underline{99.31} \\
\textbf{DECRL}          & \textbf{99.67} & \textbf{99.56} \\
\midrule 
\textbf{Improvement}    & \textbf{0.29\%} & \textbf{0.25\%} \\
\bottomrule
\end{tabular}
}
\vspace{-5pt}
\end{table}

DECRL is compared against 12 SOTA TKG representation learning approaches, categorized into shallow encoder-based approaches, i.e., TTransE \cite{leblay2018deriving} and HyTE \cite{dasgupta2018hyte}, DNN-based approaches, i.e., RE-NET \cite{jin2020}, Glean \cite{deng2020dynamic}, TeMP \cite{wu2020}, RE-GCN \cite{li2021temporal}, DACHA \cite{chen2021dacha}, and TiRGN \cite{li2022}, and derived structure-based approaches, i.e., TITer \cite{sun2021timetraveler}, EvoExplore \cite{zhang2022}, GTRL \cite{tang2023}, and DHyper \cite{tang2024}. All baselines are evaluated following consistent experimental settings with well-tuned hyper-parameters. The detailed descriptions of compared baselines can be found in Appendix \ref{descriptions_baselines}.

Tables \ref{Comparison14}, \ref{Comparison18}, and \ref{Comparisongdelt} display the MRR and Hits@1/3/10 results of event prediction on ICEWS14, ICEWS18, ICEWS14C, ICEWS18C, and GDELT datasets. “*” denotes the statistical superiority of DECRL over compared approaches based on pairwise t-tests at a 95\% confidence level. The best results are highlighted in bold, while the second-best results are underlined. Notably, DECRL consistently exhibits superior performance, yielding average improvements of 13.24\%, 12.98\%, 10.42\%, and 14.68\% in MRR and Hits@1/3/10 metrics, respectively. These findings affirm the efficacy of integrating deep evolutionary clustering for capturing the temporal evolution of high-order correlations among entities. Furthermore, it is evident that approaches leveraging derived structures generally outshine those relying on DNNs, which in turn surpass approaches employing shallow encoders.

Since most approaches do not report event prediction results on WIKI and YAGO datasets, we compare DECRL with TiRGN and DHyper, both of which provide results on these datasets, as shown in Table \ref{Comparisonwikiyago}. Furthermore, TiRGN and DHyper are the SOTA DNN-based approach and the SOTA derived structure-based approach, respectively. As a result, DECRL shows improvements of 0.29\% and 0.25\% over the runner-up, DHyper, on WIKI and YAGO datasets, respectively, underscoring its effectiveness. Detailed entity prediction results are presented in Appendix \ref{entity_prediction} due to space limitation.

\subsection{Ablation study}
\label{ablation}
To dissect the contributions of each module within DECRL, we conduct the ablation study on ICEWS14. The detailed descriptions of variants are as follows:
\begin{itemize}
    \item \textbf{DECRL-w/o-alignment}: Removing unsupervised alignment mechanism.
    \item \textbf{DECRL-w/o-fusion}: Removing fusion operation between clusters across timestamps.
    \item \textbf{DECRL-w/o-ICE}: Removing implicit correlation encoder, resulting in a fully connected graph where all edges are assigned uniform weights.
    \item \textbf{DECRL-w/o-global-graph}: Removing the guidance of the global graph on the assignment of different interaction intensities to different cluster pairs.
    \item \textbf{DECRL-w/o-$\mathcal{L}_{\mathrm{temporal}}$}: Removing the temporal smoothness loss term.
\end{itemize}

\begin{table}[]
\caption{The performance of DECRL and the variants on ICEWS14}
\renewcommand{\arraystretch}{0.9}
\setlength{\tabcolsep}{4pt}
\setlength{\abovetopsep}{0pt}
\setlength{\belowbottomsep}{0pt}
\label{tab:ablation}
\centering
{\fontsize{9}{10}\selectfont
\begin{tabular}{lllll}
\toprule
Method   &MRR &Hits@1  &Hits@3  &Hits@10\\ 
\midrule
DECRL-w/o-alignment      &59.16 (-3.5)    &43.92 (-4.8)      &67.57 (-3.0)      &92.31 (-0.7)    \\
DECRL-w/o-fusion         &57.98 (-4.3)    &41.90 (-6.8)      &66.97 (-3.6)      &92.00 (-1.0)      \\
DECRL-w/o-ICE            &60.53 (-2.8)    &46.33 (-2.4)      &68.29 (-2.3)      &91.14 (-1.9)      \\ 
DECRL-w/o-global-graph &58.67 (-3.9)    &43.45 (-5.3)      &69.46 (-1.1)      &91.07 (-2.0)       \\
DECRL-w/o-$\mathcal{L}_{\mathrm{temporal}}$ &58.74 (-3.9)  &44.43 (-4.3)  &65.33 (-5.2)  &89.85 (-3.2)\\
\midrule 
\textbf{DECRL}           &\textbf{62.61}     &\textbf{48.73}   &\textbf{70.57}    &\textbf{93.03}   \\
\bottomrule
\end{tabular}
}
\end{table}

Table \ref{tab:ablation} displays the MRR and Hits@1/3/10 results of DECRL and the variants on ICEWS14. From the ablation study, several conclusions can be drawn: Firstly, the most substantial performance degradation is observed in the DECRL-w/o-fusion variant, which indicates the effectiveness of capturing the temporal evolution of high-order correlations among entities. Moreover, the performance degradation of the DECRL-w/o-global-graph variant justifies the effectiveness of leveraging structural information from the global graph for guiding the assignment of different interaction intensities to different cluster pairs. Lastly, the performance degradation of DECRL-w/o-alignment and DECRL-w/o-$\mathcal{L}_{\mathrm{temporal}}$ variants shows the importance of preserving the temporal smoothness of clusters over successive timestamps.

\subsection{Case study}
\begin{figure}[]
\centering
\begin{subfigure}[b]{0.32\textwidth}
    \includegraphics[width=\textwidth]{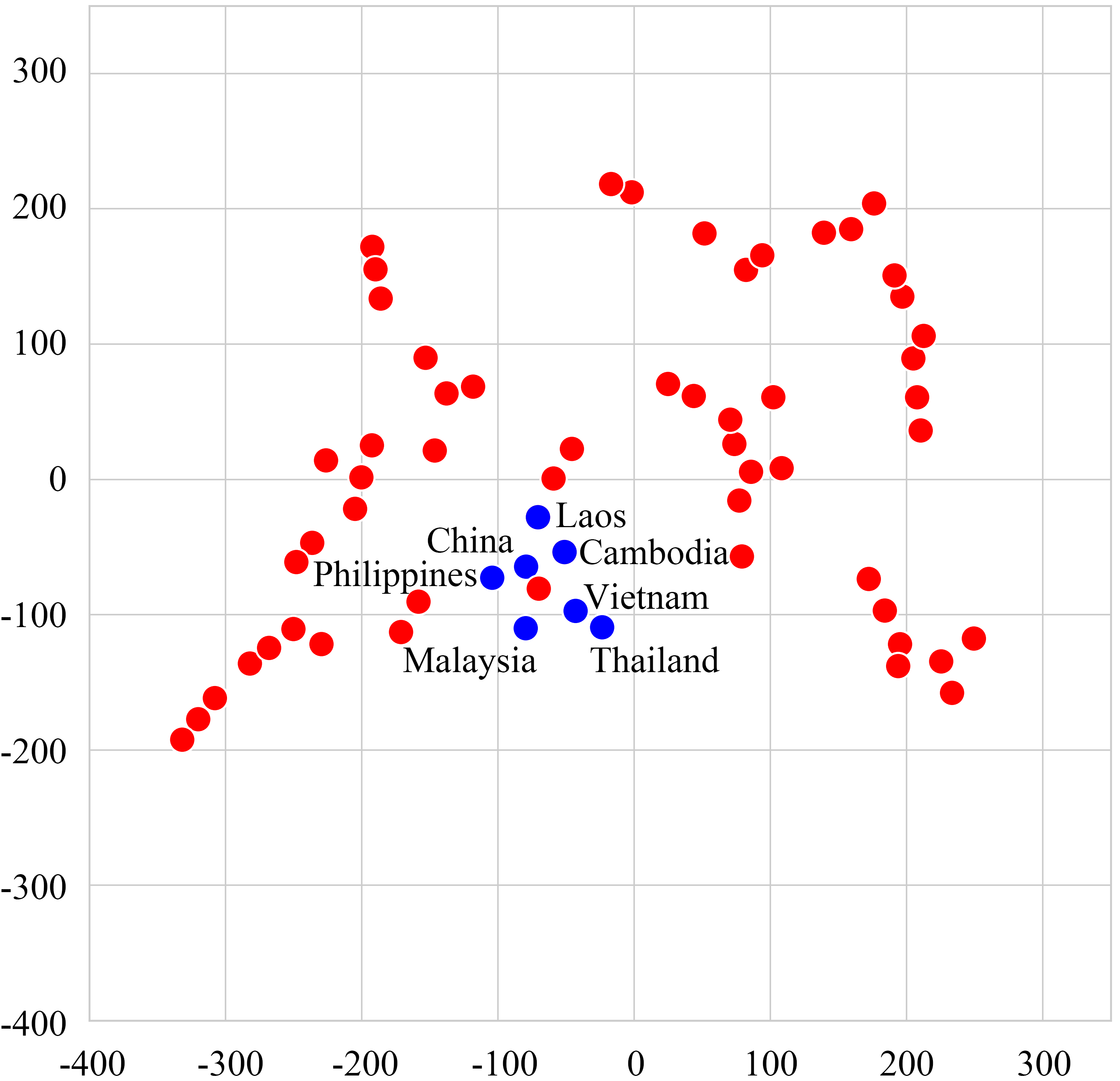}
    \caption{Middle DECRL}
    \label{fig:DECRL-middle}
\end{subfigure}
\begin{subfigure}[b]{0.32\textwidth}
    \includegraphics[width=\textwidth]{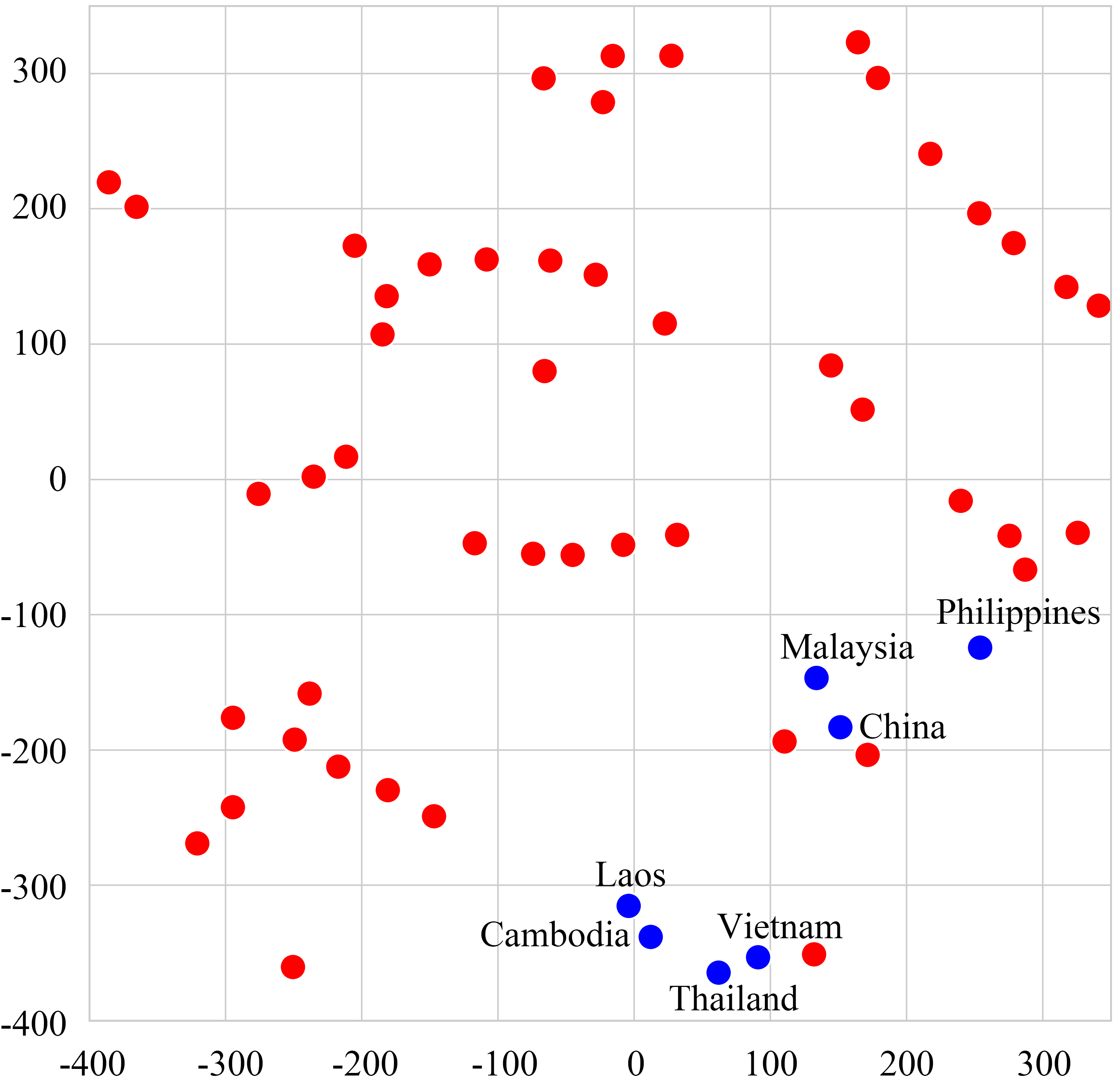}
    \caption{Middle DECRL-w/o-alignment}
    \label{fig:DECRL-wo-align-middle}
\end{subfigure}
\begin{subfigure}[b]{0.32\textwidth}
    \includegraphics[width=\textwidth]{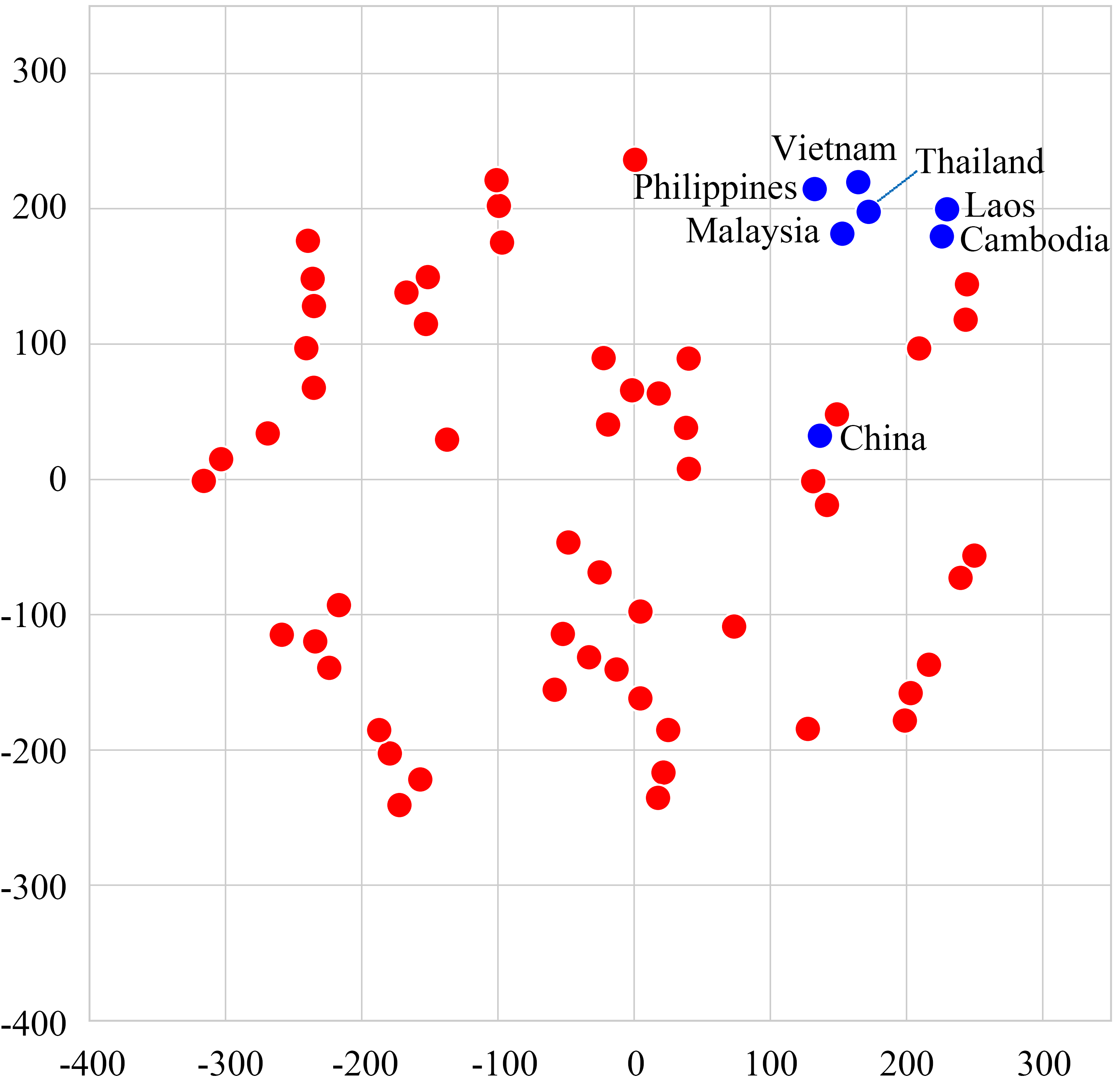}
    \caption{Middle DECRL-w/o-fusion}
    \label{fig:DECRL-wo-fusion-middle}
\end{subfigure}
\begin{subfigure}[b]{0.32\textwidth}
    \includegraphics[width=\textwidth]{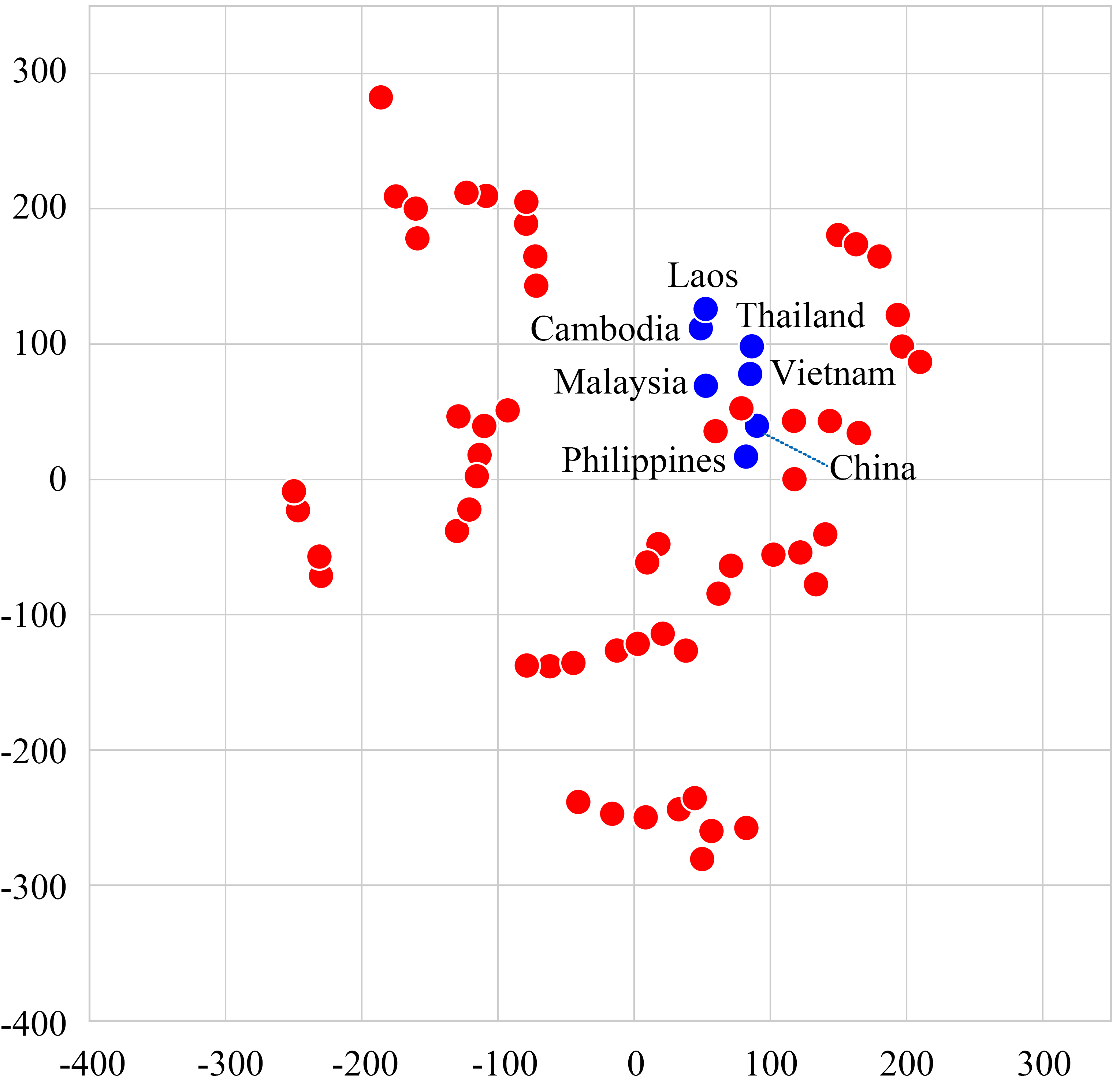}
    \caption{Final DECRL}
    \label{fig:DECRL}
\end{subfigure}
\begin{subfigure}[b]{0.32\textwidth}
    \includegraphics[width=\textwidth]{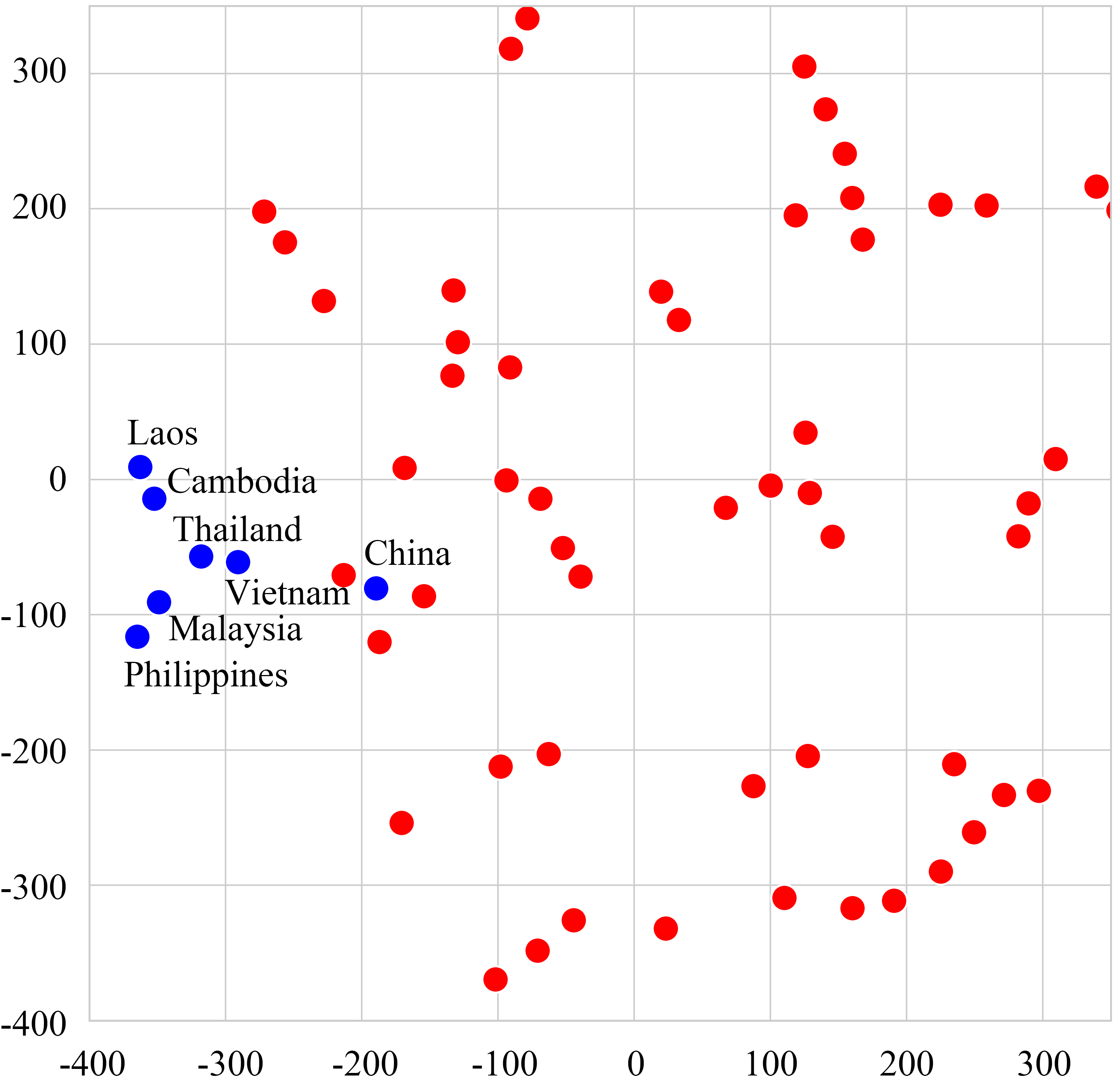}
    \caption{Final DECRL-w/o-alignment}
    \label{fig:DECRL-wo-align}
\end{subfigure}
\begin{subfigure}[b]{0.32\textwidth}
    \includegraphics[width=\textwidth]{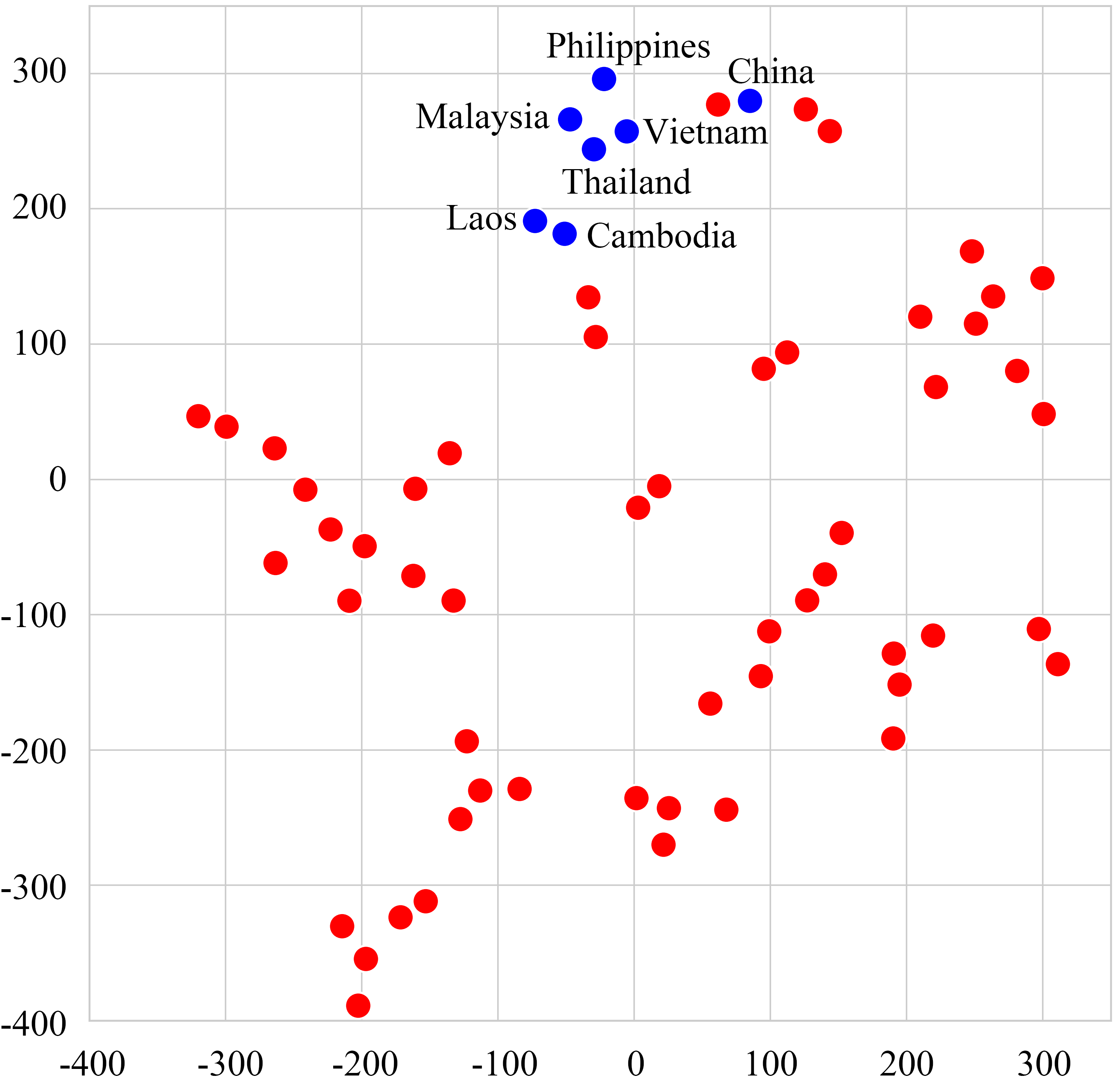}
    \caption{Final DECRL-w/o-fusion}
    \label{fig:DECRL-wo-fusion}
\end{subfigure}
\caption{The visualization of entity representations on ICEWS14C.}
\label{fig:t-SNE}
\end{figure}

To demonstrate the efficacy of DECRL, we conducted a case study with two variants: DECRL-w/o-alignment and DECRL-w/o-fusion, both of which are elaborated in Sub-section \ref{ablation}. In Figure \ref{fig:t-SNE}, we utilize t-SNE \cite{Laurens2021tsne} for the visualization of entity representations on ICEWS14C, where red dots represent individual entities in the TKGs, with their groupings indicating entity clusters. Notably, the entities marked in Figure \ref{fig:t-SNE}, i.e., China, Thailand, Vietnam, Laos, Malaysia, Philippines, and Cambodia, are countries in East Asia participating in the Belt and Road Initiative. “Middle” and “Final” denote the entity representations obtained after training at the penultimate epoch and the final epoch, respectively, for different variants.

By comparing the visualizations in the first row of sub-figures in Figure \ref{fig:t-SNE}, we observe that DECRL exhibits the best entity clustering phenomena during the mid-training phase. By comparing the visualizations in the second row of sub-figures in Figure \ref{fig:t-SNE}, Figure \ref{fig:DECRL} (Final DECRL) showcases pronounced clustering phenomena, and inter-cluster distances indicate significant differences in clusters, i.e., international political organizations. Figure \ref{fig:DECRL-wo-align} (Final DECRL-w/o-alignment) demonstrates moderate clustering phenomena among countries, but inter-cluster distances are not significant enough. This observation implies that the lack of precise one-to-one alignment may lead to proximity between different clusters. Figure \ref{fig:DECRL-wo-fusion} (Final DECRL-w/o-fusion, which only models high-order correlations without temporal evolution) exhibits increased inter-cluster distances, with less distinct clustering phenomena. The comparison between Figure \ref{fig:DECRL} (Final DECRL) and Figure \ref{fig:DECRL-wo-fusion} (Final DECRL-w/o-fusion) shows that capturing temporal evolution leads to better entity representations, as indicated by larger inter-cluster distances and tighter intra-cluster groupings. Comparing the first and third rows of Figure \ref{fig:t-SNE} further highlights the training progression, where modeling temporal evolution gradually enhances cluster separation and tightens the grouping of entities within clusters. This demonstrates that the capability of DECRL to model the temporal evolution of high-order correlations significantly enhances its ability to capture more nuanced cluster representations. Additional case study details are provided in Appendix \ref{case_study}.

\section{Conclusions and limitations}
\label{conclutions}
In this paper, a \textbf{\underline{D}}eep \textbf{\underline{E}}volutionary \textbf{\underline{C}}lustering jointed temporal knowledge graph \textbf{\underline{R}}epresentation \textbf{\underline{L}}earning approach (\textbf{DECRL}) is proposed for event prediction in TKGs, which jointly optimizes TKG representation learning with evolutionary clustering to capture the temporal evolution of high-order correlations. Comprehensive experiments are conducted on seven real-world datasets, including the comparison with baselines, ablation study, hyper-parameter sensitivity analysis, and case studies, which demonstrate the superior performance of DECRL. However, this study overlooks the continuous temporal evolution of diverse high-order correlations. Currently, the implicit correlation encoder assumes uniform correlations across all clusters, which may not reflect the complexity of real-world interactions between political organizations. In future work, we will address this issue by developing a multi relation-aware inter-cluster correlation encoder. Furthermore, DECRL currently models the temporal evolution of high-order correlations by considering only the previous and current timestamps. Future work will focus on explicitly modeling the continuous influence of high-order correlations from different past timestamps on the current timestamp.

\newpage
\section*{Acknowledgement}
This work was supported by the Science Foundation of Donghai Laboratory (Grant No. DH-2022ZY0013).

\bibliography{neurips_2024}
\bibliographystyle{plain}

\newpage
\appendix
\section{Appendix}

\subsection{Complexity analysis}
\label{complexity}
The time complexity of the relation-aware GCN is $O({N_\text{e}}{N_\text{r}}D^2)$, where $N_\text{e}$ and $N_\text{r}$ are the numbers of entities and relations, respectively. $D$ is the dimension of representations. The time complexity of the evolutionary clustering module is $O({N_\text{e}}{N_\text{c}D}+{N^3_\text{c}}+{N_\text{c}}D)$, where $N_\text{c}$ is the number of clusters. The time complexity of the cluster graph message passing module is $O(N^2_\text{c}D^2)$. The time complexity of the time residual gate is $O(D^2)$. For the attentive temporal encoder, the time complexity is $O(T^2D)$, where $T$ is the length of the history. For the event prediction module, the time complexity is $O(D)$. The total complexity of DECRL is $O(N_\text{e} N_\text{r} D^2 + N_\text{c}^3 + N_\text{c}^2 D^2 + T^2 D)$.

\subsection{Statistics of datasets}
\label{stat_dataset}

All datasets are split into training (80\%), validation (10\%), and test (10\%) sets following \cite{li2021temporal}. The statistics of these datasets are summarized in Table \ref{statistics}.

\begin{table}[]
\caption{The statistics of datasets}
\label{statistics}
\centering
{\fontsize{9}{10}\selectfont
\begin{tabular}{lllllll}
\toprule
Dataset  & \#Entity & \#Relation & \#Training set & \#Validation set & \#Test set & \#Time interval \\ 
\midrule
ICEWS14  & 7128     & 230        & 74,845        & 8,514          & 7,371    & 24 hours         \\
ICEWS14C & 205      & 171        & 35,665        & 7,369          & 7,068    & 24 hours         \\
ICEWS18  & 23,033   & 256        & 373,018       & 45,995         & 49,545   & 24 hours         \\
ICEWS18C & 208      & 164        & 34,497        & 4,412          & 4,661    & 24 hours         \\ 
GDELT    & 7,691    & 240        & 1,734,399     & 238,765        & 305,241    & 15 mins       \\
WIKI     & 12,554   & 24         & 539,286       & 67,538         & 63,110     & 1 year       \\
YAGO     & 10,623   & 10         & 161,540       & 19,523         & 20,026     & 1 year       \\
\bottomrule
\end{tabular}
}
\end{table}

\subsection{Experimental settings}
\label{exp_setting}

DECRL is implemented in Python using PyTorch and trained on one NVIDIA RTX 3080 GPU with 10GB memory. We leverage the Neural Network Intelligence (NNI) toolkit\footnote{https://github.com/microsoft/nni} to automatically identify the optimal hyper-parameter values. The search space for the number of clusters $N_\text{c}$, the number of DECRL layers $N_\text{DECRL layer}$, the length of historical windows $N_\text{historical window}$, and the value of $\lambda$ range from 1 to 20 with the step of 2, 1 to 5 with the step of 1, 1 to 14 with the step of 1, and 0.1 to 0.5 with the step of 0.1, respectively. The final hyper-parameter values are presented in Table \ref{Table2}. For NNI configurations, the maximum number of trials is set to 30, and the optimization algorithm used is the Tree-structured Parzen Estimator \cite{bergstra2015hyperopt}.

\begin{table}[]
\caption{The final choices of hyper-parameter values}
\label{Table2}
\centering
{\fontsize{9}{10}\selectfont
\begin{tabular}{llllllll}
\toprule
Hyper-parameter     & ICEWS14 & ICEWS14C & ICEWS18 & ICEWS18C  & GDELT & WIKI & YAGO\\
\midrule
$N_\text{c}$                   & 14       & 6         & 18      & 8   & 16   & 18  & 16   \\
$N_\text{DECRL layer}$               & 2        & 2         & 5       & 2    & 5    & 2  & 1 \\
$N_\text{historical window}$   & 9        & 7         & 4       & 10     & 2    & 2  & 1\\
$\lambda$                      & 0.2      & 0.2       & 0.2     & 0.2    & 0.2  & 0.2  & 0.2   \\
\bottomrule
\end{tabular}
}
\end{table}

We utilize the Adam \cite{kingma2014adam} optimizer with an initial learning rate of 0.01. The batch size is set to 16. The representation dimension is set to 200. The hidden sizes for the time residual gate and the attentive temporal encoder are both set to 200. The results reported are the averages across five independent runs. The evaluation metrics used in this paper include Mean Reciprocal Rank (MRR) and Hits@1/3/10, which represent the proportion of correct predictions ranked within the top 1, 3, and 10 positions, respectively, all expressed as percentages. Higher Hits@k and MRR scores indicate better performance.

\subsection{Description of baselines}
\label{descriptions_baselines}
\textbf{Shallow encoder-based approaches:}
\begin{itemize}
\item \textbf{TTransE} \cite{leblay2018deriving} extends the TransE by incorporating timestamps as corresponding representations.
\item \textbf{HyTE} \cite{dasgupta2018hyte} models timestamps as corresponding hyperplanes.
\end{itemize}

\textbf{DNN-based approaches:}
\begin{itemize}
\item \textbf{RE-NET} \cite{jin2020} leverages GCNs to model the influence of neighbor entities and uses RNNs to capture temporal dependencies among events.
\item \textbf{Glean} \cite{deng2020dynamic} employs CompGCN to model the influence of neighbor entities and utilizes GRUs to capture temporal dependencies among representations.
\item \textbf{TeMP} \cite{wu2020} utilizes relation-aware GCN to model the influence of neighbor entities and employs a frequency-based gating GRU to capture temporal dependencies among inactive events.
\item \textbf{RE-GCN} \cite{li2021temporal} uses relation-aware GCN to model the influence of neighbor entities and employs an autoregressive GRU to capture temporal dependencies among events.
\item \textbf{DACHA} \cite{chen2021dacha} introduces a dual graph convolution network to obtain entity representations and uses a self-attentive encoder to model temporal dependencies among relations.
\item \textbf{TiRGN} \cite{li2022} is the SOTA approach, using a multi-relational GCN to capture graph structure information and a double recurrent mechanism to model temporal dependencies.
\end{itemize}

\textbf{Derived structure-based approaches:}
\begin{itemize}
\item \textbf{TITer} \cite{sun2021timetraveler} incorporates temporal agent-based reinforcement learning to search paths and obtain entity representations via the inductive mean.
\item \textbf{EvoExplore} \cite{zhang2022} establishes dynamic communities to model the influence of neighbor entities.
\item \textbf{GTRL} \cite{tang2023} uses entity group modeling to capture the influence of distant and unreachable entities and employs GRUs to model temporal dependencies among representations.
\item \textbf {DHyper} \cite{tang2024} is the SOTA approach, which utilizes hypergraph neural networks to model high-order correlations among entities and among relations.
\end{itemize}

\subsection{Hyper-parameter sensitivity analysis}
\label{sensitivity}

\begin{figure}[]
\centering
\begin{subfigure}[b]{0.45\textwidth}
    \includegraphics[width=\textwidth]{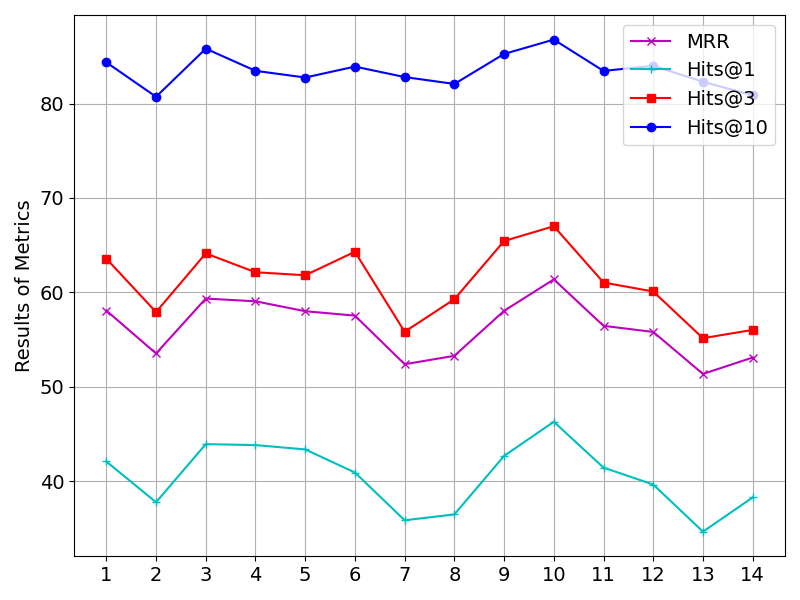}
    \caption{Hyper-parameter $N_\text{historical window}$ on ICEWS18C}
    \label{fig:historical_windows}
\end{subfigure}
\hspace{0.05\textwidth}
\begin{subfigure}[b]{0.45\textwidth}
    \includegraphics[width=\textwidth]{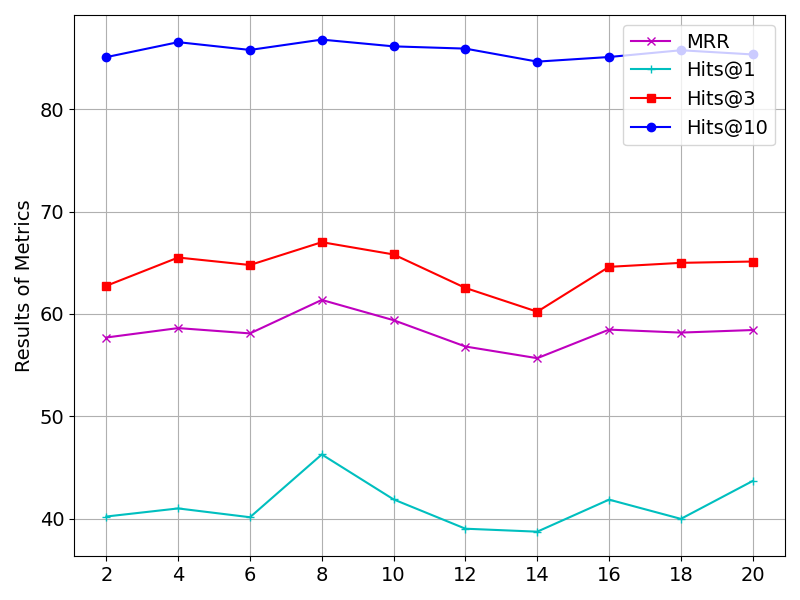}
    \caption{Hyper-parameter $N_\text{c}$ on ICEWS18C}
    \label{fig:group}
\end{subfigure}
\vspace{0.05\textwidth}
\begin{subfigure}[b]{0.45\textwidth}
    \includegraphics[width=\textwidth]{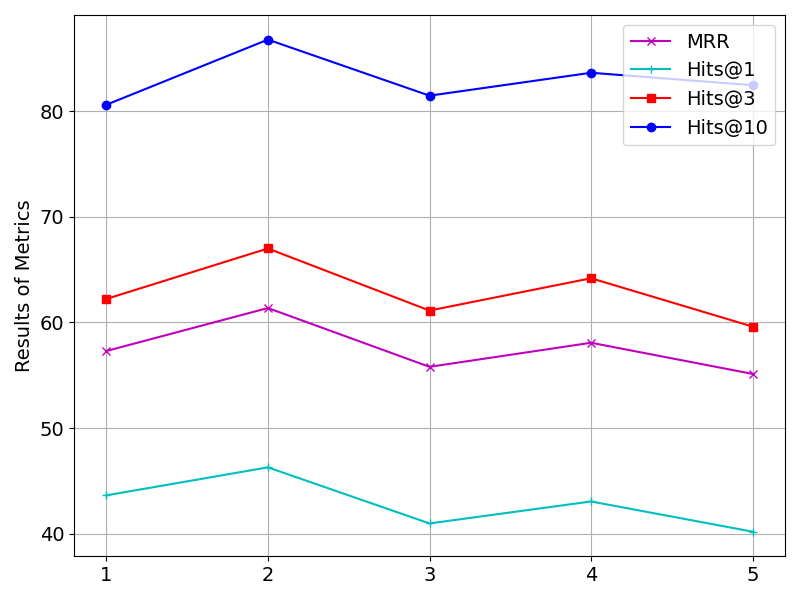}
    \caption{Hyper-parameter $N_\text{DECRL layer}$ on ICEWS18C}
    \label{fig:layer}
\end{subfigure}
\hspace{0.05\textwidth}
\begin{subfigure}[b]{0.45\textwidth}
    \includegraphics[width=\textwidth]{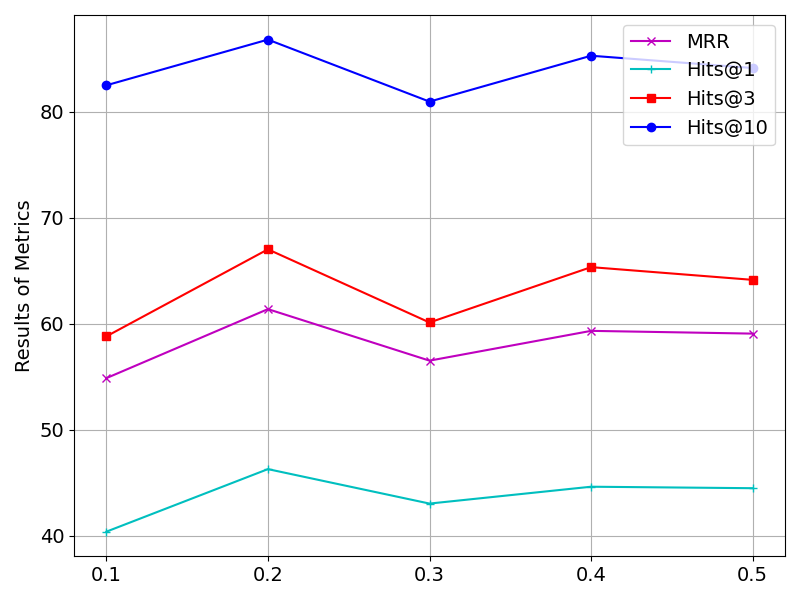}
    \caption{Hyper-parameter $\lambda$ on ICEWS18C}
    \label{fig:lambda}
\end{subfigure}
\caption{Results of hyper-parameters changes of DECRL on ICEWS18C.}
\label{fig:hyper-parameters}
\end{figure}

In this sub-section, we study the hyper-parameter sensitivity of DECRL on ICEWS18C, including the length of historical windows $N_\text{historical window}$, the number of clusters $N_\text{c}$, the number of DECRL layers $N_\text{DECRL layer}$, and the value of $\lambda$. We show the performance changes by varying the hyper-parameter values in Figure \ref{fig:hyper-parameters}.

We present the impact of $N_\text{historical window}$ in Figure \ref{fig:historical_windows}. The results indicate that as $N_\text{historical window}$ increases, the performance of DECRL gradually improves, peaking at the window length of 10. Beyond this point, performance declines rapidly, likely due to the inclusion of excessive irrelevant information in longer historical windows, which negatively impacts the performance of DECRL.

In Figure \ref{fig:group}, we present the effect of $N_\text{c}$. The performance of DECRL remains relatively stable across most metrics as $N_\text{c}$ increases. Hits@1 shows a slight initial increase until $N_\text{c} = 8$ and then stabilizes. MRR, Hits@3, and Hits@10 remain consistent, indicating that the performance of DECRL is not substantially affected by variations of $N_\text{c}$.

We show the impact of $N_\text{DECRL layer}$ and $\lambda$ in Figures \ref{fig:layer} and \ref{fig:lambda}, respectively. In Figure \ref{fig:layer}, performance metrics reach their peak when $N_\text{DECRL layer} = 2$. Beyond this point, the performance of DECRL declines, likely due to increased model complexity, which raises the risk of over-fitting. Similarly, in Figure \ref{fig:lambda}, the performance metrics peak at $\lambda = 0.2$, which suggests the need to set the appropriate trade-off between event prediction loss and temporal smoothness loss.

\subsection{Entity prediction performance}
\label{entity_prediction}

\begin{table}[]
\caption{The entity prediction performance of DECRL and the compared approaches on GDELT. “OOM” and “TLE” indicate out of memory and a single epoch exceeded 24 hours. * indicates that DECRL is statistically superior to the compared approaches according to pairwise t-test at a 95\% significance level. The best results are in bold and the second best results are underlined}
\label{Comparisoneneity}
\centering
\renewcommand{\arraystretch}{0.9}
\setlength{\tabcolsep}{4pt}
{\fontsize{9}{10}\selectfont
\begin{tabular}{lllll}
\toprule
    Approach              & MRR   & Hits@1 & Hits@3 & Hits@10 \\ \midrule
    TTransE (WWW 2018)    & 10.04* & 6.31*  & 19.23*  & 28.08*   \\
    HyTE (EMNLP 2018)     & 18.41* & 5.16*   & 21.75*  & 30.47*   \\ \midrule
    RE-NET (EMNLP 2020)   & 20.78* & 15.37*  & 23.16*  & 35.59*   \\
    Glean (KDD 2020)      & 21.14* & 19.29*  & 27.28*  & 36.37*   \\
    TeMP (EMNLP 2020)     & 36.23* & \underline{31.37}*  & 39.47*  & 48.45*   \\
    RE-GCN (SIGIR 2021)   & 21.19* & 19.53*  & 22.92*  & 35.71*   \\
    DACHA (TKDD 2022)     & 31.05* & 28.11*  & 40.12*  & 48.12*   \\
    TiRGN (IJCAI 2022)    &23.64*  &20.95*   &26.88*  & 40.26*   \\ \midrule
    TITer (EMNLP 2021)    & TLE    & TLE     & TLE     & TLE   \\
    EvoExplore (KBS 2022) & 23.61* & 17.44*  & 32.37*  & 41.47*   \\
    GTRL (TKDE 2023)      & \textbf{39.23}* & \textbf{33.37}*  & \underline{42.45}*  &\underline{51.49}*   \\
    DHyper (TOIS 2024)    &OOM     &OOM      &OOM      &OOM   \\
    \textbf{DECRL}      & \underline{37.24}  &28.60   &\textbf{46.69}   &\textbf{66.47}    \\ \midrule
    \textbf{Improvement}  &\textbf{-5.07\%} &\textbf{-14.29\%} &\textbf{9.99\%}  &\textbf{29.09\%}   \\ \bottomrule
\end{tabular}
}
\end{table}

We have conducted additional experiments to evaluate the performance of DECRL on the future entity prediction task, as shown in Table \ref{Comparisoneneity}. Although DECRL does not achieve the SOTA performance in terms of MRR and Hits@1 metrics, it achieves the best results in Hits@3 and Hits@10 metrics. It is important to note that DECRL is not specifically designed for entity prediction tasks. Nevertheless, these results demonstrate the effectiveness and robustness of DECRL, particularly in capturing a broader range of relevant entities.

\subsection{Case study}
\label{case_study}

\begin{table}[]
\centering
\caption{Top 5 relations predicted by TiRGN, DHyper, and DECRL}
\label{tab:case2}
\resizebox{\linewidth}{!}{
\begin{tabular}{m{2cm} p{5.4cm} p{5.2cm} p{5.3cm}}
\toprule
\textbf{Test sample} & \textbf{TiRGN} & \textbf{DHyper} & \textbf{DECRL} \\
\midrule
(Russia, ?, Ukraine, 2014/12/16) &
\makecell[l]{
Engage in diplomatic cooperation \\
Make empathetic comment \\
Apologize \\
Appeal for economic aid \\
Host a visit 
} & 
\makecell[l]{
\underline{Criticize or denounce} \\
Sign formal agreement \\
\underline{Reduce or break diplomatic relations} \\
Engage in negotiation \\
Praise or endorse 
} & 
\makecell[l]{
\underline{Accuse} \\
\underline{Criticize or denounce} \\
\underline{Use conventional military force} \\
Appeal for economic aid \\
\underline{Impose embargo, boycott, or sanctions} 
} \\
\midrule
(the United States, ?, Russia, 2014/12/17) &
\makecell[l]{
Make statement \\
Make a visit \\
\underline{Sign formal agreement} \\
Express intent to meet or negotiate \\
\underline{Impose embargo, boycott, or sanctions} 
} & 
\makecell[l]{
\underline{Impose embargo, boycott, or sanctions} \\
\underline{Sign formal agreement} \\
\underline{Accuse} \\
Engage in negotiation \\
\underline{Use conventional military force} 
} & 
\makecell[l]{
\underline{Disapprove} \\
\underline{Sign formal agreement} \\
\underline{Impose embargo, boycott, or sanctions} \\
\underline{Express intent to yield} \\
Use unconventional violence 
} \\
\bottomrule
\end{tabular}
}
\end{table}

Table \ref{tab:case2} presents the top 5 relations predicted by the SOTA DNN-based approach TiRGN, the SOTA derived structure-based approach DHyper, and our proposed approach DECRL for two test samples on ICEWS14C. The test samples pertain to the conflict between Russia and Ukraine that began in February 2014. During this period, Russia deployed military forces to the Crimean region of Ukraine, which resulted in widespread condemnations and sanctions from several countries, including the United States and various European nations. Correct predictions are underlined in Table \ref{tab:case2}. Compared to TiRGN and DHyper, DECRL predicts more correct relations and ranks the correct predictions higher. The results indicate that by modeling the temporal evolution of the high-order correlations among entities, DECRL can achieve more accurate prediction results.

\subsection{Societal impacts}
\label{societal_impact}
In this paper, a \textbf{\underline{D}}eep \textbf{\underline{E}}volutionary \textbf{\underline{C}}lustering jointed temporal knowledge graph \textbf{\underline{R}}epresentation \textbf{\underline{L}}earning approach (\textbf{DECRL}) is proposed for event prediction in temporal knowledge graphs, which offers more accurate and credible results. We summarize the positive and possible negative societal impacts as follows:

\textbf{Positive societal impacts:}
\begin{itemize}
\item \textbf{Optimizing social governance}: Temporal knowledge graph event prediction can help governments and relevant institutions foresee potential future events, enabling them to take preemptive measures and optimize social governance.
\item \textbf{Supporting business decisions}: Companies can use the prediction results for market analysis and business decisions, enhancing their competitiveness.
\item \textbf{Enhancing public safety}: By predicting potential threats and dangerous events, relevant departments can allocate resources in advance to ensure public safety.
\item \textbf {Advancing academic research}: This research can promote progress in the fields of temporal knowledge graphs and event prediction, providing new directions and methods for academic studies.
\end{itemize}

\textbf{Negative societal impacts:}
\begin{itemize}
\item \textbf{Misuse risks}: Accurate predictive technology might be exploited by malicious actors to forecast and manipulate events.
\end{itemize}

\end{document}